\documentclass{article}
\usepackage{amssymb}
\usepackage[hang]{footmisc}

\usepackage[dvipsnames]{xcolor}
\usepackage{bbm}
\usepackage{amsmath}
\usepackage{graphics}
\usepackage{multirow}
\usepackage{amssymb}
\usepackage{booktabs}
\usepackage{tikz}
\usepackage{xspace}
\usepackage{booktabs,caption}
\usepackage[flushleft]{threeparttable}
\usepackage{subcaption}
\usepackage{algorithm}
\usepackage{algorithmic}
\usepackage{wrapfig}
\usepackage{wrapfig}
\usepackage{float}

\definecolor{Green4}{RGB}{1,50,32}

\newcommand{\greencheck}{{\color{Green4} \checkmark}}
\newcommand{\redcross}{{\color{red}$\times$}}

\newcommand{\figref}[1]{Fig.~\ref{#1}}
\newcommand{\secref}[1]{Sec.~\ref{#1}}

\newcommand{\tabref}[1]{Tab.~\ref{#1}}
\newcommand{\suppl}{\textcolor{black}{SupMat}\xspace}

\newcommand{\etal}{et al.}

\usepackage{cuted}

\usepackage[preprint]{corl_2025} %

\title{Robust Dexterous Grasping of General Objects}

\newcommand\blfootnote[1]{%
  \begingroup
  \renewcommand\thefootnote{}\footnote{#1}%
  \addtocounter{footnote}{-1}%
  \endgroup
}

\author{
Hui Zhang{\rm \! \footnotesize\textsuperscript{1,*}} \;\; 
Zijian Wu{\rm \! \footnotesize\textsuperscript{2,*}} \;\;
Linyi Huang{\rm \! \footnotesize\textsuperscript{2}} \;\;
Sammy Christen{\rm \! \footnotesize\textsuperscript{1}} \;\;
Jie Song{\rm \! \footnotesize\textsuperscript{2,3}}
}

\begin{document}
\maketitle

\vspace{-3em}
\begin{center}
\textsuperscript{1}ETH Zurich, Switzerland \;\; \textsuperscript{2} HKUST (Guangzhou), China \;\; \textsuperscript{3} HKUST, Hong Kong (China)
\end{center}

\begin{figure}[!h]
	\centering
	\includegraphics[width=0.99\linewidth]{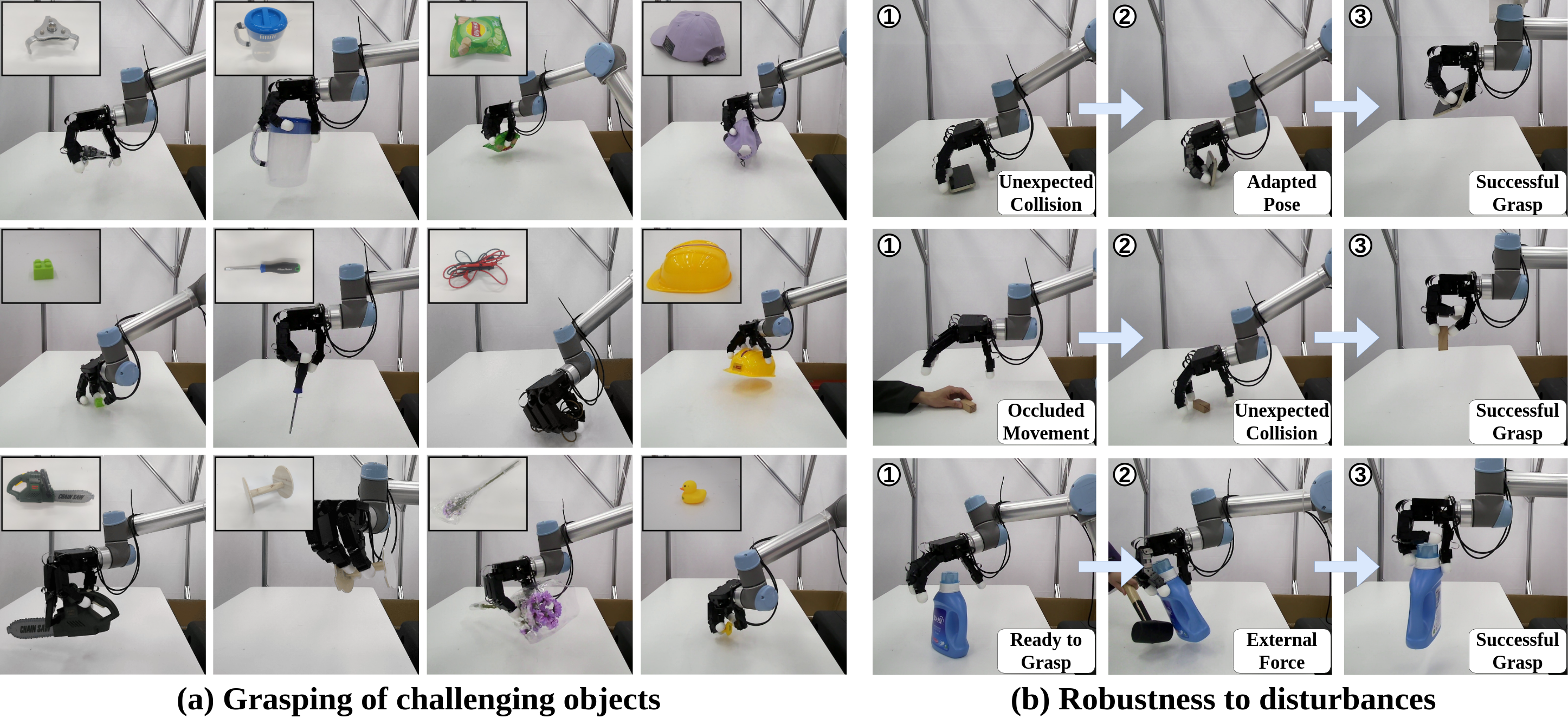}
	\caption{Our method can robustly grasp 500+ unseen objects with various shapes, materials, and masses, including challenging thin, small, heavy, deformable, or transparent objects (a). It can keep robust grasping under unexpected collision, imprecise object observation, or external forces (b).}
\vspace{-4mm}
	\label{fig:adaptive}
\end{figure}

\begin{abstract}
    The ability to robustly grasp a variety of objects is essential for dexterous robots. In this paper, we present a framework for zero-shot dynamic dexterous grasping using single-view visual inputs, designed to be resilient to various disturbances. Our approach utilizes a hand-centric object shape representation based on dynamic distance vectors between finger joints and object surfaces. This representation captures the local shape around potential contact regions rather than focusing on detailed global object geometry, thereby enhancing generalization to shape variations and uncertainties. To address perception limitations, we integrate a privileged teacher policy with a mixed curriculum learning approach, allowing the student policy to effectively distill grasping capabilities and explore for adaptation to disturbances. Trained in simulation, our method achieves success rates of 97.0\% across 247,786 simulated objects and 94.6\% across 512 real objects, demonstrating remarkable generalization. Quantitative and qualitative results validate the robustness of our policy against various disturbances.     Additional results and supplementary videos are provided on our \href{https://zdchan.github.io/Robust_DexGrasp/}{project website}.
\end{abstract}


\section{Introduction}

\vspace{-1.5em}
\blfootnote{
\vspace{0.3em}
\textsuperscript{$\ast$}Equal Contribution. 
\vspace{-1.5em}
}

Grasping is a fundamental capability in robotic manipulation, forming the foundation for higher-level tasks such as pick-and-place, articulating hinged objects, tool use, and in-hand manipulation. 
Dexterous hands, with their high degrees of freedom, enable more flexible and adaptive grasping. Realizing this potential, however, requires precise closed-loop control with real-time feedback to ensure robustness under environmental variations and disturbances.
Previous research on dexterous grasping has mainly focused on settings with fully observable or pre-scanned known objects~\cite{huang2024fungrasp, wang2022dexgraspnet}, human-to-robot demonstration transfer~\cite{chen2022dextransfer, chen2024object, li2025maniptrans}, or the execution of predicted static grasping poses or fingertip positions~\cite{attarian2023geometry, chen2024springgrasp, zhang2024dexgraspnet2}. However, these approaches often rely on pre-scanned objects or object-specific human demonstrations, which limit their efficient deployment to novel objects. Moreover, grasping strategies based on static pose execution limit real-time adaptability, resulting in reduced robustness to disturbances, especially when predefined poses become infeasible.

In this paper, we propose a framework for zero-shot dexterous grasping of diverse objects using single-view visual inputs. Similar to human capabilities, our system can adapt in real-time to disturbances such as unexpected collisions arise from noisy observations or external perturbations.
Our approach utilizes a hand-centric object shape representation with dynamic distance vectors between finger joints and object surfaces, focusing on potential interactions rather than global shapes. This design improves robustness to shape variations and perceptual uncertainty caused by limited viewpoints. Robust adaptation to disturbances requires continuous observation feedback, yet perception during grasping is often constrained by occlusion and the lack of tactile sensing. We address this challenge through a two-step training paradigm: first training a teacher policy using privileged visual-tactile information, then employing a mixed curriculum learning approach to train a student policy. The curriculum starts with imitation learning to efficiently distill the core grasping behaviors of the teacher policy, and gradually transitions to reinforcement learning to promote exploration of adaptation to disturbances. During student policy training, observation noise and dynamic randomization are introduced to simulate real-world perturbations and encourage robust adaptation.

We conduct comprehensive experiments to evaluate the generalization and robustness of our method. Our approach shows strong generalizability in grasping a wide range of novel objects with random poses on a tabletop.
Trained entirely in simulation,
it achieves success rates of \textbf{97.0\%} on \textbf{247,786} simulated objects and \textbf{94.6\%} on \textbf{512} \textbf{real-world} objects.
Furthermore, our method facilitates real-time adaptation during grasping, exhibiting superior robustness against disturbances such as unexpected external forces.
An ablation study further confirms the importance of key components, validating their contributions to overall performance.
 
In summary, our contributions are 1) 
A robust dexterous grasping framework designed for general objects, equipped with real-time adaptation capabilities to handle disturbances effectively.
2) A sparse hand-centric object shape representation tailored for real-world dexterous grasping, which captures interaction potential and facilitates robustness against shape variations and perceptual uncertainty.
3) 
A mixed curriculum learning method that integrates imitation learning for efficient grasping behavior distillation and reinforcement learning for adaptive motion exploration, under limited perception.
4) 
Our method, trained in simulation, demonstrates exceptional generalization across 247,786 unseen objects in simulation and 512 unseen objects on a real robot. It also exhibits robust adaptive motions in response to disturbances such as external forces. 
Additionally, we qualitatively showcase the practical applications enabled by our robust grasping capability.

\section{Related Work}

We list our differences with existing dexterous grasping methods in \tabref{tab:related_work} for a better comparison. 

\textbf{Pose-based Dexterous Grasping}.
Dexterous grasping is a long-standing research topic~\cite{ferrari1992planning, Kim2014catching, Takahashi08adaptive, wang2025unigrasptransformer}. Traditional methods typically predict contact points or fingertip positions by optimizing analytical metrics for stable grasping, such as the differentiable approximations of shape closure~\cite{wang2022dexgraspnet} and force closure~\cite{liu2021synthesizing} metrics. These works mostly require accurate object models~\cite{Ciocarlie2007DexterousGV, ferrari1992planning, Miller2004graspit, Roa2015quality} which limits their generalization ability in real-world deployment. 
Recently, some works have explored dealing with object shape uncertainty with compliant control algorithms~\cite{chen2024springgrasp, Kazemi-RSS-12, LI2016ras, Pfanne2020Impedance, Puhlmann2022RBO}. They usually utilize analytical dynamic models to calculate joint torque commands according to static target poses, limiting their robustness to unmodeled disturbances such as inaccurate joint actuator models or external forces. 
Instead of optimizing grasping poses or fingertip positions, some other methods learn to predict grasping poses from datasets~\cite{zhang2024multi, zhang2024dexgraspnet2}. 
Overall, these methods usually focus on generating static grasping poses and executing them in an open-loop manner without adaptation, which limits hand dexterity and robustness to external disturbances, as the hand cannot adaptively change poses when the object moves unexpectedly. In contrast, our method predicts real-time joint actions according to current status, leading to adaptive motions to disturbances and more robust grasping.

\textbf{Dynamic Dexterous Grasping}.
Rather than predicting static grasping poses, recent works explore real-time joint action prediction for dynamic dexterous grasping. Many of these approaches leverage human demonstrations~\cite{chen2022dextransfer, chen2024object, li2025maniptrans, ye2023learning, okami2024, liu2025dextrack} or robot demonstrations~\cite{yang2024ace, cheng2024tv, jiang2024dexmimicen, fang2025anydexgrasp, zhong2025dexgraspvla} to learn such actions. However, collecting real-world data remains costly, limiting generalization to out-of-distribution scenarios.
Reinforcement learning (RL) has shown promise in handling disturbances across environments, particularly in robot locomotion~\cite{Choi2023sr, gu2024advancing, miki2022learning}. Through exploration in simulation under disturbances, policies can learn dynamic actions and perform real-time adaptation. However, due to complex hand-object interactions that are challenging to explore and simulate, current RL-based dexterous grasping methods are often limited in specific settings (e.g., category-level generalization~\cite{dexpoint}) and suffer from sim-to-real gaps.
Overall, most RL-based dynamic dexterous grasping methods are still confined to simulation~\cite{christen2022dgrasp, Wan_2023_ICCV, yuan2024cross, zhang2024graspxl, zhang2024artigrasp}.
Some RL-based methods leverage human data to simplify exploration and enable deployment on real robots. For example, ~\cite{Agarwal2023functional, lum2024dextrahg, singh2024dextrahrgbvisuomotorpoliciesgrasp} simplify RL exploration with a lower-dimensional action space using PCA components derived from human grasping data, limiting hand dexterity as admitted in their papers.~\cite{Agarwal2023functional} relies on a multi-camera system with only category-level generalization, while~\cite{lum2024dextrahg, singh2024dextrahrgbvisuomotorpoliciesgrasp} use analytical dynamic models for control commands, limiting adaptation to unmodeled disturbances.~\citet{huang2024fungrasp} leverage human grasping poses while relying on known object meshes.
Overall, requirements for specific human data, accurate dynamic models, and known objects limit their potential to scale up or adapt to unmodeled disturbances. 
Our method, in contrast, can be deployed on real robots with a single camera, achieving zero-shot generalization to 512 unseen real-world objects without any human data or known object meshes, while performing real-time proprioception-based adaptive motions to external disturbances.

\begin{table}
\caption{Comparison with other dexterous grasping works}
\vspace{1mm}
\centering
\resizebox{0.82\columnwidth}{!}{
\begin{tabular}{l|ccccc}
   \toprule
    
\multirow{2}{*}{Method} & Single-view  & Dynamic  & Zero-shot & Unseen Object & Unseen Object \\
                        & Observation   & Grasping  & Generalization   & Number (Sim)    & Number (Real) \\
    \midrule
    GraspXL \cite{zhang2024graspxl} & \redcross   & \greencheck & \greencheck   & 503,409 & -  \\
    Agarwal \etal \cite{Agarwal2023functional} & \redcross  & \greencheck & \redcross  & 2 & 6   \\
    SpringGrasp \cite{chen2024springgrasp} & \greencheck & \redcross   & \greencheck   & - & 14  \\
    DextrAH-G \cite{lum2024dextrahg} & \greencheck & \greencheck   & \greencheck   & - & 30  \\
    DexGraspNet2.0 \cite{zhang2024dexgraspnet2} & \greencheck  & \redcross   & \greencheck   & 1319 & 32  \\
    Ours           & \greencheck & \greencheck & \greencheck  & 247,786  & 512 \\
   \bottomrule
\end{tabular}
}
\label{tab:related_work}
 \vspace{-5mm}
\end{table}

\section{Method}

\begin{figure}[t]
  \centering
  \includegraphics[width=0.99\textwidth]{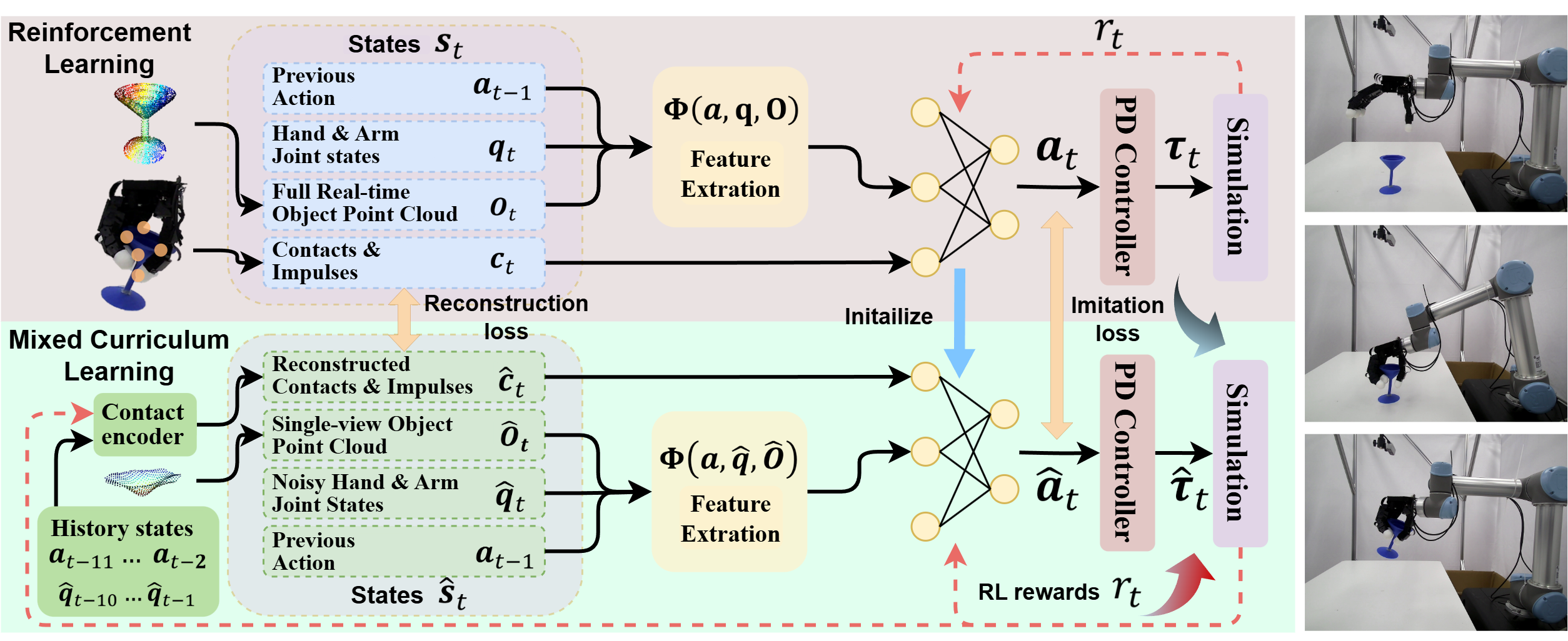}
  \caption{\textbf{Framework overview}. We first train a teacher policy with privileged visual-tactile perception driven by RL. Then we train a student policy with single-view object point clouds and noisy proprioception, driven by our mixed curriculum learning method, which starts with IL for efficient teacher policy distillation, and gradually transitions to RL for exploration under disturbances. 
  }
  \vspace{-5mm}
\label{fig:pipeline}
\end{figure}

We aim to tackle dexterous grasping of various unseen objects using single-view visual perception. Given a single-view object point cloud, we control a robotic arm with a dexterous hand to grasp the object while adapting to disturbances. \figref{fig:pipeline} illustrates our pipeline. We first train a teacher policy using reinforcement learning (RL) with access to real-time, fully observable object point clouds and hand-object contacts and impulses, noted as a visual-tactile policy.
Then we train a student policy with perception available on real robots, including single-view point clouds and noisy joint states without tactile perception.
The student policy training is driven by a mixed curriculum learning method, starting with imitation learning (IL) to efficiently distill the teacher's grasping capabilities, then gradually transitioning to RL to explore for adaptive motions to noises and disturbances.
Both policies output target hand and arm joint positions for low-level PD controllers.

\begin{wrapfigure}{r}{0.23\columnwidth}
    \vspace{-17mm}
	\centering
	\includegraphics[width=0.23\columnwidth]{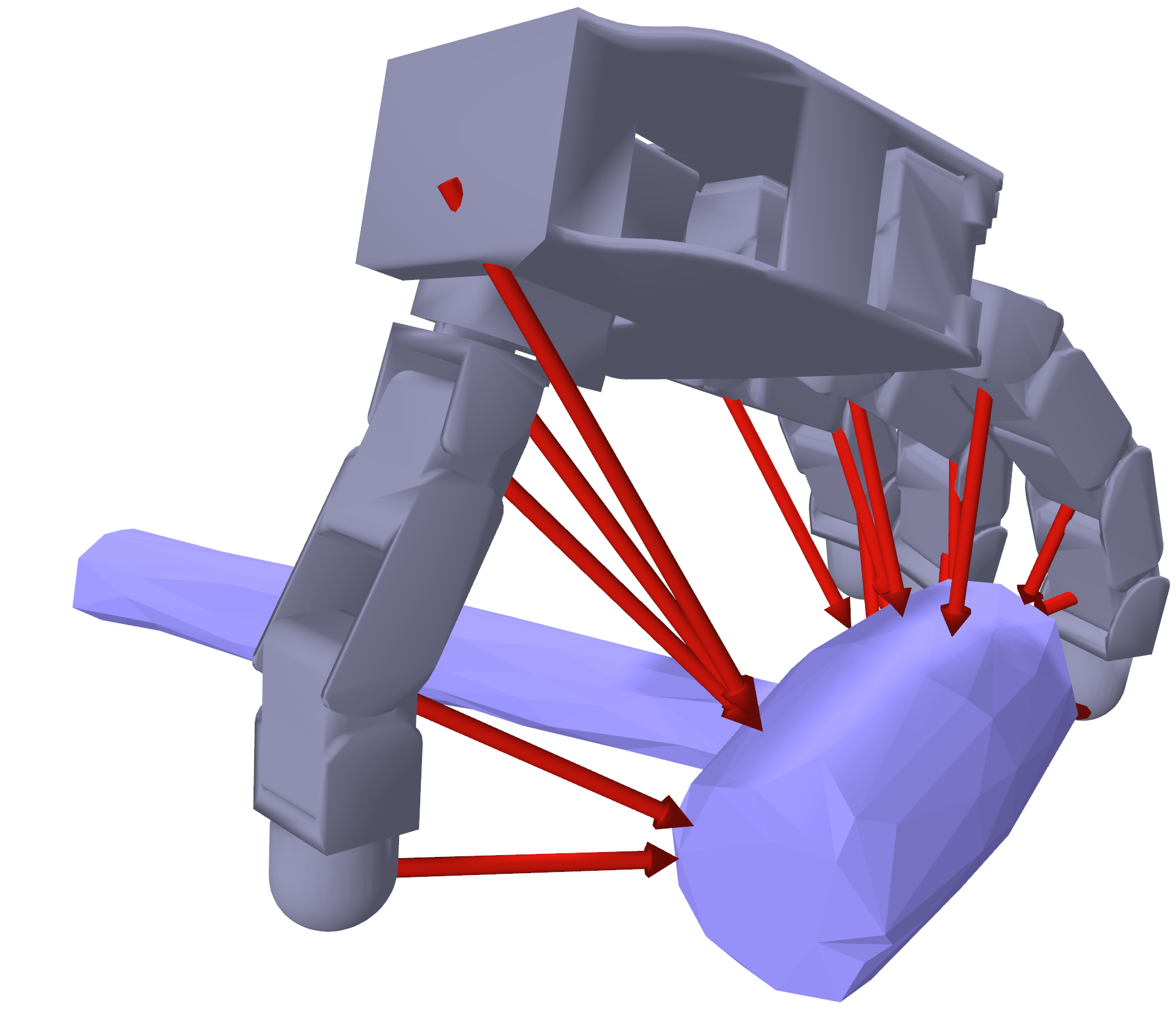}
	\caption{Hand-centric Shape Representation}
	\label{fig:shape}
    \vspace{-5mm}
\end{wrapfigure}

\subsection{Hand-centric Object Shape Representation}
Effectively capturing shape features that generalize for grasping diverse objects remains a fundamental challenge, especially under shape uncertainty arising from limited viewpoints. To address this, we introduce a sparse hand-centric object representation. As shown in \figref{fig:shape}, we construct a 51-dimensional vector by concatenating the distance vectors from each finger joint (including the 
wrist and fingertips) to their nearest points from the object point clouds. This compact representation efficiently encodes interaction-relevant local shape features instead of fine-grained global geometry, which filters out irrelevant shape variations in non-contact areas, facilitates generalization to novel object shapes, and improves robustness under perceptual uncertainty. Its effectiveness is verified in \secref{sec:ablation}.

\subsection{Visual-tactile Policy Training}
We first train a policy with RL to control the robot arm and dexterous hand, which has access to real-time full object point clouds and hand-object binary contacts and impulses of each finger link.

\textbf{Observation}.
The observation space of the policy is represented with $\textbf{s}_t = (\textbf{a}_{t-1}, \textbf{q}_t, \textbf{O}_t, \textbf{c}_t)$, where $t$ is the current time step, $\textbf{a}_{t-1}$ is the action of the previous step, $\textbf{q}_t$ is the current arm and finger joint angles, $\textbf{O}_t$ is the real-time fully observable object point cloud, $\textbf{c}_t=\{\textbf{b}_t, \textbf{f}_t\}$ includes the binary contact states $\textbf{b}_t$ and impulses $\textbf{f}_t$ of each finger link with the object. We omit the notation ``$_t$'' for simplicity in this section. Given $\textbf{a}$, $\textbf{q}$ and $\textbf{O}$, we extract the following features with a function $\Phi(\textbf{a}, \textbf{q}, \textbf{O}) = (\textbf{d}, \textbf{h}, \textbf{T}, \Delta\textbf{q})$. Specifically, $\textbf{d}$ is the vector from each finger joint to the closest point of $\textbf{O}$.
$\textbf{h}$ is the vertical distance of each arm and hand link to the table. $\textbf{T}$ is the wrist orientation and position. $\Delta\textbf{q}$ is the tracking error, which means the difference between current joint angles $\textbf{q}_t$ and previous action $\textbf{a}_{t-1}$. The extracted features are then fed to the policy together with $\textbf{q}$ and $\textbf{c}$. 

\textbf{Reward Design}.
We design the reward function to encourage robust and safe grasping with the formulation 
$r = r_{\text{dis}} + r_{\text{contact}} + r_{\text{height}} + r_{\text{reg}}$.

The distance reward $r_{\text{dis}}$ encourages the hand to approach the object by penalizing the link-object distances with 
$\scalebox{0.9}{$r_{\text{dis}} = -\sum_{i=1}^L w_{i}^{d}\cdot||\textbf{d}_i||^2$}$,
where \scalebox{0.9}{$w_{i}^{d}$} is the weights and $L$ is the number of hand links. 

The contact reward $r_{\text{contact}}$ promotes desired hand-object contacts while penalizing undesired contacts, including self-collision, robot-table collision, and arm-object collision, 
with the formulation 
\scalebox{0.9}{$r_{\text{contact}} = \sum_{i=1}^{L} b_i (w_{i}^{cd} + w_{i}^{fd} f^o_i) - \sum_{j=1}^{L+M} b_j (w_{j}^{cu} + w_{j}^{fu} f^u_j)$}. \scalebox{0.9}{$w_{i}^{cd}, w_{j}^{cu}$, $w_{i}^{fd}, w_{j}^{fu}$} are the weights for desired and undesired binary contact states and impulses, $b_i$ and $b_j$ are the desired binary contact states of the $i_{th}$ hand link and undesired binary contact states of the $j_{th}$ hand or arm link, $f^o_j$ and $f^u_j$ are the magnitude of desired hand impulses on the $i_{th}$ link and undesired hand or arm impulses on the $j_{th}$ link, and $M$ is the number of arm links.

The height reward $r_{\text{height}}$ promotes table collision avoidance by discouraging the robot links from excessively approaching the table, achieved by penalizing link-table distances when they are smaller than 2 cm with
\scalebox{0.9}{$r_{\text{height}} = \sum_{i=1}^{L+M} w_{i}^{h}\cdot \text{log}(\text{min}\{h_i, 0.02\})$},
where $w_{i}^{h}$ is the weight and $h_i$ is the vertical distance to the table of the $i_{th}$ link.

The regularization reward $r_{\text{reg}}$ penalizes unnecessary object motion and extreme robot movements with
\scalebox{0.9}{$r_{\text{reg}} = w_h||\dot{\textbf{T}}_h||^2 + w_o||\dot{\textbf{T}}_o||^2 + w_l||\textbf{l}_o|| + w_q||\dot{\textbf{q}}_a||^2$},
where $w_h, w_o, w_l, w_q$ are the weights, $\dot{\textbf{T}}_h$ and $\dot{\textbf{T}}_o$ are the linear and angular velocities of the hand and object, $\textbf{l}_o$ is the object displacement, and $\dot{\textbf{q}}_a$ is the arm joint velocities.

All weight values can be found in \suppl.

\subsection{Student Policy Training with Limited Perception}
With the teacher policy trained using real-time visual-tactile perception, we further train a student policy with 
single-view object point clouds and noisy proprioception without tactile information. We randomize the friction coefficients and proprioception of robot joint angles when training the student policy. Besides, we also randomize the PD gains of the low-level joint controllers to simulate unstable hardware actuator stiffness and damping, potentially caused by factors like overheating (especially for finger joint actuators).
All randomization parameters can be found in \suppl.

\textbf{Observation}.
The visual-tactile policy requires real-time fully observable object point clouds $\textbf{O}_t$ and contact states $\textbf{c}_t$, which are privileged information inaccessible on the real robots. For hardware deployment, the student policy should utilize the 
single-view object point cloud $\hat{\textbf{O}}_t$ and noisy joint state proprioception $\hat{\textbf{q}}_t$. Specifically, we utilize an LSTM-based encoder to reconstruct the contact states $\hat{\textbf{c}}_t=\{\hat{\textbf{b}}_t, \hat{\textbf{f}}_t\}$ from joint state and action histories. Intuitively, the actions correspond to the joint actuator torques, while the misalignment between the actuator torques and joint state changes can indicate external forces induced by contacts.

\textbf{Mixed Curriculum Learning}.
While the teacher policy focuses on stable grasping with visual-tactile perception, the student policy must learn to grasp and adapt to disturbances with limited perception. To deal with this challenge, we propose a mixed curriculum learning approach. The training begins with IL using two losses: a contact reconstruction loss $L_{re} = w_{re}(|| \hat{\textbf{b}}_t - \textbf{b}_t ||^{2} + || \hat{\textbf{f}}_t - \textbf{f}_t ||^{2})$ to reconstruct contacts $\textbf{c}_t$, and an action imitation loss $L_{act} = w_{act}|| \hat{\textbf{a}}_t - \textbf{a}_t||^{2}$ to mimic the teacher policy actions. This can help the student policy efficiently distill the grasping capability of the teacher policy. Training then gradually transitions to RL by decreasing $w_{act}$ with a factor $\lambda$ and increasing RL rewards by $1-\lambda$ (keeping $w_{re}$ fixed). This transition encourages the student policy to keep exploration for adaptation to disturbances from observation noises and actuator inaccuracies. The student policy network is initialized with teacher policy weights to accelerate training.

\section{Experiments}

\begin{wrapfigure}{r}{0.38\columnwidth}
    \vspace{-12mm}
	\centering
	\includegraphics[width=0.38\columnwidth]{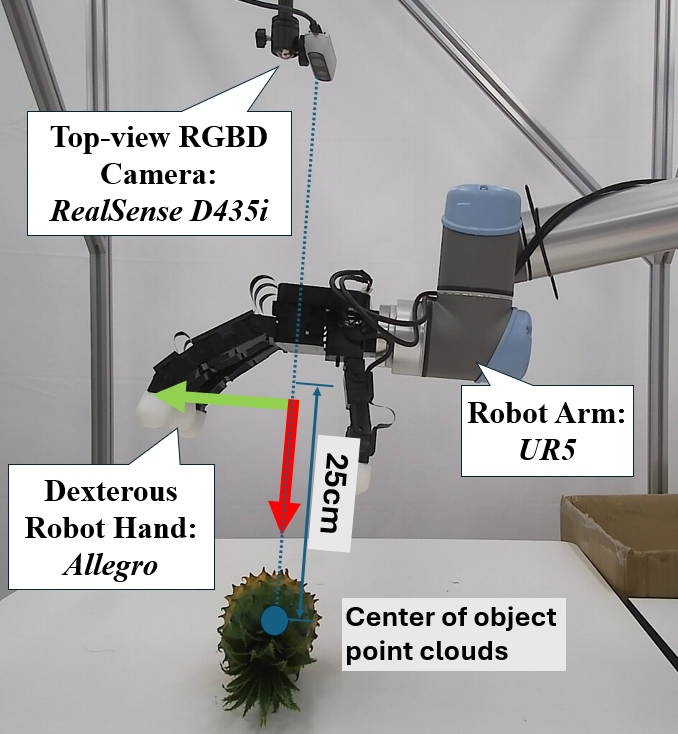}
	\caption{Hardware setup}
	\label{fig:hardware}
    \vspace{-8mm}
\end{wrapfigure}

\subsection{Experimental Setup}

\textbf{Hardware Setup}. 
The robot platform used in this work comprises a UR5~\cite{UR5} robot arm paired with an Allegro~\cite{Allegro} dexterous robot hand. We utilize a top-view RealSense D435i camera. An overview of the hardware setup is shown in \figref{fig:hardware}.
The hand and arm are initialized with a partially opened hand heading to object centers from 25cm away, and enclosing the objects with the fingers. The policy runs at 5 Hz as a high-level controller, while the low-level PD controllers for the hand and arm operate at 100 Hz. More details are explained in \suppl.

\textbf{Training Details}. 
Our policy is trained in Raisim~\cite{hwangbo2018raisim} simulation, using 35 objects including 3D assets from~\cite{ycb} and scanned objects from~\cite{huang2024fungrasp}. 

\textbf{Metric}. 
Since our focus is on robust grasping, we use the grasping success rate (Suc. Rate) as our evaluation metric. Specifically, a grasp is considered successful only if the object can be lifted to a height of 0.1 meters and remains stable without falling for at least 5 seconds.

\subsection{Large Scale Evaluation}
To demonstrate the strong generalization ability of our method, we evaluate it using 512 real objects and 247,786 simulated objects, which are all unseen during training.

\begin{table}[t]
    \caption{Large-scale evaluation (Sim)}
\vspace{1mm}
    \centering
    \resizebox{0.52\columnwidth}{!}{
    \begin{tabular}{l|cccc}
        \toprule
        Size & Small & Medium & Large & Total \\
        \midrule
        Object Number & 38,493  & 100,435 & 108,858  & 247,786 \\
        \midrule
        Success Rate & 0.949 & 0.972  & 0.976   &  0.970 \\
        \bottomrule
    \end{tabular}}
        \vspace{-3mm}
    \label{tab:large_scale_sim}
\end{table}

\textbf{Simulation Evaluation}. 
We begin by evaluating our method on objects from the Objaverse~\cite{objaverse} dataset. Following ~\cite{zhang2024graspxl}, we scale the objects into three sizes: small, medium, and large. Objects that are infeasible to grasp from a tabletop setting (i.e., diameter $<$ 1cm, height $<$ 2cm, or width $>$ 15cm) are excluded, resulting in a final test set of 247,786 objects. Preprocessing details are provided in \suppl. Objects are randomly placed on the table for grasping. As shown in \tabref{tab:large_scale_sim}, our method achieves an overall success rate of 0.970, with consistently high performance across different object sizes. Notably, while grasping smaller objects is more challenging due to the relatively large link dimensions of the Allegro hand, our method still achieves a 0.949 success rate on small objects. These results highlight the strong generalization capability of our approach across diverse object shapes and scales, which we attribute to the proposed sparse hand-centric shape representation.

\begin{table}[t]
    \caption{Large-scale evaluation (Real)}
    \vspace{1mm}
    \centering
       \resizebox{\columnwidth}{!}{
    \begin{tabular}{c|c|c|c|c|c|c|c}
        \toprule
        Category & Suc. Rate & Category & Suc. Rate & Category & Suc. Rate & Category & Suc. Rate  \\
        \midrule
        Picnic Models & 0.902 & Building Blocks  & 0.963 & Fruit \& Vegetable Models & 0.962    & Tool Models  & 0.875 \\
        Animal Models & 0.907 & Toy Cars & 0.979 & Wooden Models & 0.940 & Snacks &  0.974\\
        Bottles \& Boxes & 0.970 & Real Tools  & 0.893 & Deformable Objects & 0.957           & Other Daily Objects  & 0.971 \\
        \midrule
        Average & \multicolumn{7}{c}{0.946}  \\
        \bottomrule
    \end{tabular}}
            \vspace{-2mm}
    \label{tab:large_scale_real}
\end{table}

\textbf{Hardware Evaluation}. 
We further assess the effectiveness of our method in grasping a wide range of novel real-world objects. To this end, we collect a set of 512 objects from different categories, with variations in shape, weight (7 g to 610 g), material (e.g., plastic, styrofoam, rubber, wood, glass, paper, metal, cloth, sponge), and size (from 3.5×3.5×1.5 cm to 40×20×13 cm). Detailed object specifications are provided in \suppl. Each object is evaluated in three random table-top poses.
The average success rate per category is reported in \tabref{tab:large_scale_real}. Using only single-view depth observation without any prior knowledge of the objects, our method achieves an overall success rate of 0.946. Notably, despite being trained exclusively on rigid objects in simulation, the learned policy generalizes to deformable objects, demonstrating robust sim-to-real transfer. We attribute this generalization to the reconstructed tactile information along finger links, which enables adaptive finger closure until a stable force-closure grasp is formed. Most failure cases are caused by two factors: noisy point cloud observations for thin or small objects, and insufficient finger torques for heavy or slippery objects due to hardware limitations of the actuators.

\subsection{Method Comparison}

\label{sec:eval}
One key advantage of RL-based grasping is its real-time adaptation ability, especially under disturbances. To verify this, we first evaluate our method in real-world using 30 3D-printed ShapeNet~\cite{shapenet2015} objects (shown in \suppl), compared with the following baselines:
1) a naive controller that moves the hand above the objects and gradually closes the fingers. 
2) DexGraspNet~\cite{wang2022dexgraspnet} (with an extra table collision penalty term), with the optimized grasping poses executed by PD controllers. 
We also evaluate our method with two state-of-the-art methods using their settings and objects: 
3) SpringGrasp~\cite{chen2024springgrasp}, an optimization-based compliant grasping method. 
4) DextrAH-G~\cite{lum2024dextrahg}, an RL-based dynamic grasping method. 
To further verify our robustness to disturbances, we compare our method with DexGraspNet under 
external forces in both simulation and real-world settings. 
Each object is tested across five random poses. More details of the baselines are in \suppl. 

\begin{table}[t]
    \caption{Method comparison (Real)}
    \vspace{1mm}
    \centering
    \resizebox{0.99\columnwidth}{!}{
    \begin{tabular}{cc|cc|cc|cc}
        \toprule
        Method & Suc. Rate & Method & Suc. Rate  & Method & Suc. Rate & Method & Suc. Rate   \\
        \midrule
         Naive baseline & 0.633 & DexGraspNet~\cite{wang2022dexgraspnet} & 0.607 & SpringGrasp \cite{chen2024springgrasp} & 0.771 & DextrAH-G \cite{lum2024dextrahg} & 0.927  \\
        \midrule
         Ours  &  0.920  &  Ours  &  0.920 & Ours  &  0.957 &  Ours  &  0.964  \\
        \bottomrule
    \end{tabular}}
    \vspace{-3mm}
    \label{tab:compare}
\end{table}

\textbf{Real-world Comparison without External Disturbances}. 
We first compare the real-world performance of different methods without explicit external disturbances. It is worth noting, however, that hardware experiments inherently involve disturbances due to noisy joint states, object observations, and actuator inaccuracies (e.g., caused by overheating). Results are shown in \tabref{tab:compare}. Our method consistently outperforms all baselines, regardless of whether being evaluated on our or their settings and object sets, highlighting the enhanced robustness of our RL-based policy.
We argue that robust grasping requires real-time adaptive control over finger selection, timing, and joint torques to accommodate diverse object properties, which are effectively learnt by our RL-based policy. In contrast, the naive baseline and DexGraspNet often fail due to limited adaptability, particularly with thin, smooth, or round objects, and may knock objects over due to misaligned finger contact.
SpringGrasp optimizes fingertip positions, initial hand poses, and controller gains, allowing the hand to grasp the object compliantly. However, it still lacks the capability to adjust grasping poses according to real-time status, leading to lower performance compared with our method. Among all baselines, DextrAH-G achieves the closest performance to ours, benefiting from its RL-based dynamic grasping. However, as acknowledged in their paper, its PCA-constrained action space restricts dexterity, and its model-based table collision avoidance reduces performance on thin objects. In contrast, our method preserves the full degree-of-freedom of the action space and actively handles rather than rigidly avoids table contacts, leading to improved performance for thin objects as \figref{fig:adaptive} shows.

\begin{table}[t]
    \caption{Robustness to external forces (Sim \& Real)}
    \vspace{1mm}
    \centering
       \resizebox{0.9\columnwidth}{!}{
    \begin{tabular}{c|cc|c|cc}
       \toprule
        Method &  No Disturbance  & 2.5 N Force & Method & No Disturbance  & 2.5 N Force \\
        \midrule
        DexGraspNet \cite{wang2022dexgraspnet} (Sim)  &  0.667  & 0.553  & DexGraspNet \cite{wang2022dexgraspnet} (Real)  &  0.607 & 0.480\\
        \midrule
        Ours (Sim) & 0.953  & 0.920  &   Ours (Real) &  0.920   & 0.840\\
        \bottomrule
    \end{tabular}}
    \label{tab:force}
    \vspace{-2mm}
\end{table}

\textbf{Robustness to External Forces}. 
We further evaluate the robustness of our method against external forces applied to objects after they have been grasped, comparing the performance with DexGraspNet. 
In simulation, we apply an external force of 2.5 N with random directions at a random point on the object. In the hardware experiment, we place a 250 g weight at a randomly selected location on the object to introduce a 2.5 N downward force as a controlled external disturbance.
As shown in \tabref{tab:force}, our method achieves superior success rates with smaller performance degradation under disturbances in both simulation and real-world environments, 
indicating enhanced robustness against external forces. We argue the robustness comes from our policy's ability to consistently generate stable force-closure grasps and adapt in real time to in-hand object slips. This capability is reinforced through randomized PD gains and friction coefficients during training, requiring the student policy to maintain stable grasps under dynamic disturbances. In contrast, the baseline struggles to consistently achieve robust force-closure grasps and perform real-time adaptation, especially for thin objects.
We observe that heavier objects are more susceptible to external forces for both methods, especially in hardware experiments due to the physical torque limits of the finger joint actuators.

\subsection{Ablation and Analysis}
\label{sec:ablation}

We conduct an ablation study to assess the impact of individual components in both simulation and the real world. We use the same 30 ShapeNet objects as in \secref{sec:eval}, with 5 random poses per object.

\textbf{Simulation Ablation}.
In simulation, we first evaluate the effectiveness of our mixed curriculum learning approach by comparing against the following variants that train the student policy: 1) without the RL rewards (W.o. RL rewards), 2) without the action imitation loss (W.o. IL loss), 3) using a fixed ratio for the RL rewards and imitation loss (W.o. Curriculum). Secondly, to examine the necessity of a privileged teacher policy, we include a variant that directly trains the student policy from scratch using only RL rewards and contact reconstruction loss (W.o. Priv. learning). Finally, we compare our student policy with the teacher policy to validate its effectiveness (Teacher policy).

\begin{table}[t]
    \caption{Ablation (Sim \& Real)}
    \vspace{1mm}
    \centering
    \resizebox{0.68\columnwidth}{!}{
    \begin{tabular}{c|c|c|c|c}
        \toprule
        & Setting & Suc. Rate & Setting & Success Rate   \\
        \midrule
        \multirow{3}{*}{Sim} &  W.o. RL rewards & 0.907      & W.o. IL loss &  0.933  \\ 
        & W.o. Curriculum  & 0.913   &  W.o. Priv. learning  &  0.773  \\
        \cmidrule(lr){2-5}
        &  Ours & 0.953   &  Teacher policy   & 0.960 \\
        \midrule
        Real & Ours & 0.920 & Full Point Cloud & 0.933 \\
        \bottomrule
    \end{tabular}}
    \vspace{-2mm}
    \label{tab:ablation}
\end{table}

The results are presented in \tabref{tab:ablation}. Compared to our full method, the variants without RL rewards and curriculum exhibit a similar decline in performance, underscoring the necessity of the exploration enabled by RL. The marginally lower success rate achieved without IL loss indicates its contribution to student policy training. Overall, starting with a larger IL loss facilitates a rapid distillation of the grasping capabilities from the teacher policy, while subsequently increasing RL rewards encourages effective exploration for adaptation to disturbances. This illustrates the effectiveness of our mixed curriculum learning approach.
Training the student policy from scratch (W.o. Priv. learning) yields substantially lower performance and increased sensitivity to hyperparameters, underscoring the importance of the privileged teacher in providing effective initialization and supervision during early training stages.
The comparable success rates between our student and teacher policies further validate the effectiveness of our mixed curriculum learning method.

\textbf{Real-world Ablation}.
We further compare our original setting with a variant that utilizes real-time, fully observable object point clouds in the real world (Full Point Cloud), which we get from FoundationPose~\cite{wen2024foundationpose} along with known object meshes.
In contrast, our approach uses
single-view object point clouds captured by the camera without object meshes, which can be directly applied to novel objects for zero-shot generalization. The results are shown in \tabref{tab:ablation}, 
where the highly comparable performances between the two settings highlight the effectiveness of our sparse object-centric representation in extracting meaningful shape features from single-view object observations.

\subsection{Qualitative Results}
We qualitatively demonstrate the generalization and robustness of our method in \figref{fig:adaptive}. Our policy can deal with various challenging objects, and adaptively adjust grasping poses in response to unexpected collisions caused by noisy joint states, inaccurate object observations, or actuator imprecision. It also effectively compensates for disturbances and maintains stable grasps under significant external forces. Additional qualitative results are provided in the Suppl. Video, where we further showcase various applications enabled by our robust grasping capability with different modules integrated, such as grasping in cluttered scenes with a segmentation model~\cite{kirillov2023sam}, task-driven grasping with a VLM planner~\cite{bai2023qwen}, and grasping moving objects with an object tracking module~\cite{wen2021bundletrack}.

\section{Conclusion}

In this paper, we propose an RL-based framework for robust dexterous grasping from single-view perception. Our method exhibits strong generalization capability, achieving success rates of 94.6\% on 512 real-world objects and 97.0\% on 247,786 simulated objects. Compared to existing methods, our RL-based dynamic grasping approach exhibits superior robustness and adaptability, especially under randomly applied external forces and inherent internal disturbances such as observation noises or actuator inaccuracies, which can lead to unexpected collisions and require pose adaptation.
We show that a sparse, hand-centric object shape representation enables effective generalization across diverse object geometries under perceptual uncertainty. Furthermore, our mixed curriculum learning framework facilitates the policy to learn real-time adaptive motions under varying disturbances with limited perception. These components together constitute a robust, low-level grasping controller, providing the foundation for diverse downstream applications as demonstrated in our Suppl. Video.

\clearpage
\section{Limitations}

Although our method demonstrates strong generalization and robustness across a wide range of objects, it still has several limitations that we plan to address in future work.
First, despite its capability to grasp objects with diverse sizes and physical properties, the large link dimensions of the Allegro Hand make it difficult to grasp very small objects (e.g., with diameters less than 1.5 cm). Solving this limitation relies on a smaller dexterous robot hand with a similar scale to the human hand.
Second, although our policy can grasp moving objects as demonstrated in the Suppl. Video, it struggles in highly dynamic environments such as catching a flying object. This limitation could potentially be addressed by incorporating an explicit object motion prediction module.
Finally, our framework focuses specifically on robust grasping without addressing non-grasping interactions like pushing objects, which are also essential in many manipulation tasks. Such interactions may be enabled in the future through task-specific reward designs and policy training.

\bibliography{example}  %

\begin{thebibliography}{52}
\providecommand{\natexlab}[1]{#1}
\providecommand{\url}[1]{\texttt{#1}}
\expandafter\ifx\csname urlstyle\endcsname\relax
  \providecommand{\doi}[1]{doi: #1}\else
  \providecommand{\doi}{doi: \begingroup \urlstyle{rm}\Url}\fi

\bibitem[Huang et~al.(2024)Huang, Zhang, Wu, Christen, and Song]{huang2024fungrasp}
L.~Huang, H.~Zhang, Z.~Wu, S.~Christen, and J.~Song.
\newblock \href{https://arxiv.org/abs/2411.16755}{{FunGrasp}: Functional Grasping for Diverse Dexterous Hands}.
\newblock \emph{arXiv preprint arXiv:2411.16755}, 2024.

\bibitem[Wang et~al.(2023)Wang, Zhang, Chen, Xu, Li, Liu, and Wang]{wang2022dexgraspnet}
R.~Wang, J.~Zhang, J.~Chen, Y.~Xu, P.~Li, T.~Liu, and H.~Wang.
\newblock \href{https://ieeexplore.ieee.org/stamp/stamp.jsp?arnumber=10160982}{DexGraspNet: A Large-Scale Robotic Dexterous Grasp Dataset for General Objects Based on Simulation}.
\newblock \emph{{International Conference on Robotics and Automation (ICRA)}}, 2023.

\bibitem[Chen et~al.(2022)Chen, Van~Wyk, Chao, Yang, Mousavian, Gupta, and Fox]{chen2022dextransfer}
Z.~Q. Chen, K.~Van~Wyk, Y.-W. Chao, W.~Yang, A.~Mousavian, A.~Gupta, and D.~Fox.
\newblock \href{https://arxiv.org/abs/2209.14284}{Dextransfer: Real world multi-fingered dexterous grasping with minimal human demonstrations}.
\newblock \emph{arXiv preprint arXiv:2209.14284}, 2022.

\bibitem[Chen et~al.(2024)Chen, Wang, Yang, and Liu]{chen2024object}
Y.~Chen, C.~Wang, Y.~Yang, and K.~Liu.
\newblock \href{https://arxiv.org/abs/2411.04005}{Object-Centric Dexterous Manipulation from Human Motion Data}.
\newblock In \emph{Conference on Robot Learning (CoRL)}, 2024.

\bibitem[Li et~al.(2025)Li, Li, Liu, Li, and Huang]{li2025maniptrans}
K.~Li, P.~Li, T.~Liu, Y.~Li, and S.~Huang.
\newblock \href{https://arxiv.org/abs/2503.21860}{{ManipTrans}: Efficient Dexterous Bimanual Manipulation Transfer via Residual Learning}.
\newblock In \emph{{Computer Vision and Pattern Recognition (CVPR)}}, 2025.

\bibitem[Attarian et~al.(2023)Attarian, Asif, Liu, Hari, Garg, Gilitschenski, and Tompson]{attarian2023geometry}
M.~Attarian, M.~A. Asif, J.~Liu, R.~Hari, A.~Garg, I.~Gilitschenski, and J.~Tompson.
\newblock \href{https://openreview.net/forum?id=oyWkrG-LD5}{Geometry Matching for Multi-Embodiment Grasping}.
\newblock In \emph{Conference on Robot Learning (CoRL)}, 2023.

\bibitem[Chen et~al.(2024)Chen, Bohg, and Liu]{chen2024springgrasp}
S.~Chen, J.~Bohg, and C.~K. Liu.
\newblock \href{https://arxiv.org/abs/2404.13532}{SpringGrasp: An optimization pipeline for robust and compliant dexterous pre-grasp synthesis}.
\newblock \emph{{Robotics: Science and Systems (RSS)}}, 2024.

\bibitem[Zhang et~al.(2024)Zhang, Liu, Li, Yu, Geng, Ding, Chen, and Wang]{zhang2024dexgraspnet2}
J.~Zhang, H.~Liu, D.~Li, X.~Yu, H.~Geng, Y.~Ding, J.~Chen, and H.~Wang.
\newblock \href{https://arxiv.org/abs/2410.23004}{{DexGraspNet 2.0}: Learning Generative Dexterous Grasping in Large-scale Synthetic Cluttered Scenes}.
\newblock \emph{Conference on Robot Learning (CoRL)}, 2024.

\bibitem[Ferrari and Canny(1992)]{ferrari1992planning}
C.~Ferrari and J.~Canny.
\newblock \href{https://ieeexplore.ieee.org/document/219918}{Planning optimal grasps}.
\newblock \emph{{International Conference on Robotics and Automation (ICRA)}}, 1992.

\bibitem[Kim et~al.(2014)Kim, Shukla, and Billard]{Kim2014catching}
S.~Kim, A.~Shukla, and A.~Billard.
\newblock \href{https://ieeexplore.ieee.org/document/6810147}{Catching Objects in Flight}.
\newblock \emph{{Transactions on Robotics (T-RO)}}, 2014.

\bibitem[Takahashi et~al.(2008)Takahashi, Tsuboi, Kishida, Kawanami, Shimizu, Iribe, Fukushima, and Fujita]{Takahashi08adaptive}
T.~Takahashi, T.~Tsuboi, T.~Kishida, Y.~Kawanami, S.~Shimizu, M.~Iribe, T.~Fukushima, and M.~Fujita.
\newblock \href{https://ieeexplore.ieee.org/document/4543219}{Adaptive grasping by multi fingered hand with tactile sensor based on robust force and position control}, 2008.

\bibitem[Wang et~al.(2025)Wang, Wei, Zhou, Chen, Luo, Yi, Zhang, Liang, Xu, Lu, Yang, and Guo]{wang2025unigrasptransformer}
W.~Wang, F.~Wei, L.~Zhou, X.~Chen, L.~Luo, X.~Yi, Y.~Zhang, Y.~Liang, C.~Xu, Y.~Lu, J.~Yang, and B.~Guo.
\newblock \href{https://arxiv.org/abs/2412.02699}{{UniGraspTransformer}: Simplified Policy Distillation for Scalable Dexterous Robotic Grasping}.
\newblock In \emph{{Computer Vision and Pattern Recognition (CVPR)}}, 2025.

\bibitem[Liu et~al.(2021)Liu, Liu, Jiao, Zhu, and Zhu]{liu2021synthesizing}
T.~Liu, Z.~Liu, Z.~Jiao, Y.~Zhu, and S.-C. Zhu.
\newblock \href{https://ieeexplore.ieee.org/stamp/stamp.jsp?arnumber=9619920}{Synthesizing diverse and physically stable grasps with arbitrary hand structures using differentiable force closure estimator}.
\newblock \emph{{Robotics and Automation Letters (RA-L)}}, 2021.

\bibitem[Ciocarlie et~al.(2007)Ciocarlie, Goldfeder, and Allen]{Ciocarlie2007DexterousGV}
M.~T. Ciocarlie, C.~Goldfeder, and P.~K. Allen.
\newblock \href{https://www.semanticscholar.org/paper/Dexterous-Grasping-via-Eigengrasps-%3A-A-Approach-to-Ciocarlie-Goldfeder/6df8653fc8758d0d1534207e34000a5189abf212}{Dexterous Grasping via Eigengrasps: A Low-dimensional Approach to a High-complexity Problem}.
\newblock 2007.

\bibitem[Miller and Allen(2004)]{Miller2004graspit}
A.~Miller and P.~Allen.
\newblock \href{https://ieeexplore.ieee.org/document/1371616/authors#authors}{Graspit! A versatile simulator for robotic grasping}.
\newblock \emph{IEEE Robotics \& Automation Magazine}, 2004.

\bibitem[Roa and Su\'{a}rez(2015)]{Roa2015quality}
M.~A. Roa and R.~Su\'{a}rez.
\newblock \href{https://dl.acm.org/doi/10.1007/s10514-014-9402-3}{Grasp quality measures: review and performance}.
\newblock \emph{Auton. Robots}, 2015.

\bibitem[Kazemi et~al.(2012)Kazemi, Valois, Bagnell, and Pollard]{Kazemi-RSS-12}
M.~Kazemi, J.-S. Valois, J.~A. Bagnell, and N.~Pollard.
\newblock \href{https://www.ri.cmu.edu/pub_files/2012/7/RSS2012_Final_Compliant_Grasp.pdf}{Robust Object Grasping using Force Compliant Motion Primitives}.
\newblock \emph{{Robotics: Science and Systems (RSS)}}, 2012.

\bibitem[Li et~al.(2016)Li, Hang, Kragic, and Billard]{LI2016ras}
M.~Li, K.~Hang, D.~Kragic, and A.~Billard.
\newblock \href{https://ieeexplore.ieee.org/document/6810147}{Dexterous grasping under shape uncertainty}.
\newblock \emph{Robotics and Autonomous Systems}, 2016.

\bibitem[Pfanne et~al.(2020)Pfanne, Chalon, Stulp, Ritter, and Albu-Schäffer]{Pfanne2020Impedance}
M.~Pfanne, M.~Chalon, F.~Stulp, H.~Ritter, and A.~Albu-Schäffer.
\newblock \href{https://ieeexplore.ieee.org/document/9001211}{Object-Level Impedance Control for Dexterous In-Hand Manipulation}.
\newblock \emph{{Robotics and Automation Letters (RA-L)}}, 2020.

\bibitem[Puhlmann et~al.(2022)Puhlmann, Harris, and Brock]{Puhlmann2022RBO}
S.~Puhlmann, J.~Harris, and O.~Brock.
\newblock \href{https://ieeexplore.ieee.org/document/9761831}{{RBO} Hand 3: A Platform for Soft Dexterous Manipulation}.
\newblock \emph{{Transactions on Robotics (T-RO)}}, 2022.

\bibitem[Zhang et~al.(2024)Zhang, Bai, Huang, Chen, and Zhang]{zhang2024multi}
L.~Zhang, K.~Bai, G.~Huang, Z.~Chen, and J.~Zhang.
\newblock \href{https://arxiv.org/abs/2404.08844}{Multi-fingered Robotic Hand Grasping in Cluttered Environments through Hand-object Contact Semantic Mapping}.
\newblock \emph{arXiv preprint arXiv:2404.08844}, 2024.

\bibitem[Ye et~al.(2023)Ye, Wang, Huang, Qin, and Wang]{ye2023learning}
J.~Ye, J.~Wang, B.~Huang, Y.~Qin, and X.~Wang.
\newblock \href{https://ieeexplore.ieee.org/document/10081001}{Learning continuous grasping function with a dexterous hand from human demonstrations}.
\newblock \emph{{Robotics and Automation Letters (RA-L)}}, 2023.

\bibitem[Li et~al.(2024)Li, Zhu, Xie, Jiang, Seo, Pavlakos, and Zhu]{okami2024}
J.~Li, Y.~Zhu, Y.~Xie, Z.~Jiang, M.~Seo, G.~Pavlakos, and Y.~Zhu.
\newblock \href{https://arxiv.org/abs/2410.11792}{{OKAMI}: Teaching Humanoid Robots Manipulation Skills through Single Video Imitation}.
\newblock \emph{Conference on Robot Learning (CoRL)}, 2024.

\bibitem[Liu et~al.(2025)Liu, Adalibieke, Han, Qin, and Yi]{liu2025dextrack}
X.~Liu, J.~Adalibieke, Q.~Han, Y.~Qin, and L.~Yi.
\newblock \href{https://arxiv.org/abs/2502.09614}{{DexTrack}: Towards Generalizable Neural Tracking Control for Dexterous Manipulation from Human References}.
\newblock In \emph{{International Conference on Learning Representations (ICLR)}}, 2025.

\bibitem[Yang et~al.(2024)Yang, Liu, Qin, Runyu, Jialong, Cheng, Yang, Yi, and Wang]{yang2024ace}
S.~Yang, M.~Liu, Y.~Qin, D.~Runyu, L.~Jialong, X.~Cheng, R.~Yang, S.~Yi, and X.~Wang.
\newblock \href{https://arxiv.org/abs/2408.11805}{{ACE}: A Cross-platfrom Visual-Exoskeletons for Low-Cost Dexterous Teleoperation}.
\newblock \emph{arXiv preprint arXiv:2408.11805}, 2024.

\bibitem[Cheng et~al.(2024)Cheng, Li, Yang, Yang, and Wang]{cheng2024tv}
X.~Cheng, J.~Li, S.~Yang, G.~Yang, and X.~Wang.
\newblock \href{https://arxiv.org/abs/2407.01512}{{Open-TeleVision}: Teleoperation with Immersive Active Visual Feedback}.
\newblock \emph{Conference on Robot Learning (CoRL)}, 2024.

\bibitem[Jiang et~al.(2024)Jiang, Xie, Lin, Xu, Wan, Mandlekar, Fan, and Zhu]{jiang2024dexmimicen}
Z.~Jiang, Y.~Xie, K.~Lin, Z.~Xu, W.~Wan, A.~Mandlekar, L.~Fan, and Y.~Zhu.
\newblock \href{https://arxiv.org/abs/2410.24185}{{DexMimicGen}: Automated Data Generation for Bimanual Dexterous Manipulation via Imitation Learning}.
\newblock \emph{arXiv preprint arXiv:2410.24185}, 2024.

\bibitem[Fang et~al.(2025)Fang, Yan, Tang, Fang, Wang, and Lu]{fang2025anydexgrasp}
H.-S. Fang, H.~Yan, Z.~Tang, H.~Fang, C.~Wang, and C.~Lu.
\newblock \href{https://arxiv.org/abs/2502.16420}{{AnyDexGrasp}: General Dexterous Grasping for Different Hands with Human-level Learning Efficiency}.
\newblock \emph{arXiv preprint arXiv:2502.16420}, 2025.

\bibitem[Zhong et~al.(2025)Zhong, Huang, Li, Zhang, Liang, Yang, and Chen]{zhong2025dexgraspvla}
Y.~Zhong, X.~Huang, R.~Li, C.~Zhang, Y.~Liang, Y.~Yang, and Y.~Chen.
\newblock \href{https://arxiv.org/abs/2502.20900}{{DexGraspVLA}: A Vision-Language-Action Framework Towards General Dexterous Grasping}.
\newblock \emph{arXiv preprint arXiv:2502.20900}, 2025.

\bibitem[Choi et~al.(2023)Choi, Ji, Park, Kim, Mun, Lee, and Hwangbo]{Choi2023sr}
S.~Choi, G.~Ji, J.~Park, H.~Kim, J.~Mun, J.~H. Lee, and J.~Hwangbo.
\newblock \href{https://www.science.org/doi/10.1126/scirobotics.ade2256}{Learning quadrupedal locomotion on deformable terrain}.
\newblock \emph{Science Robotics}, 2023.

\bibitem[Gu et~al.(2024)Gu, Wang, Zhu, Shi, Guo, Liu, and Chen]{gu2024advancing}
X.~Gu, Y.-J. Wang, X.~Zhu, C.~Shi, Y.~Guo, Y.~Liu, and J.~Chen.
\newblock \href{https://arxiv.org/abs/2408.14472}{Advancing Humanoid Locomotion: Mastering Challenging Terrains with Denoising World Model Learning}.
\newblock \emph{{Robotics: Science and Systems (RSS)}}, 2024.

\bibitem[Miki et~al.(2022)Miki, Lee, Hwangbo, Wellhausen, Koltun, and Hutter]{miki2022learning}
T.~Miki, J.~Lee, J.~Hwangbo, L.~Wellhausen, V.~Koltun, and M.~Hutter.
\newblock \href{https://www.science.org/doi/10.1126/scirobotics.abk2822}{Learning robust perceptive locomotion for quadrupedal robots in the wild}.
\newblock \emph{Science Robotics}, 2022.

\bibitem[Qin et~al.(2022)Qin, Huang, Yin, Su, and Wang]{dexpoint}
Y.~Qin, B.~Huang, Z.-H. Yin, H.~Su, and X.~Wang.
\newblock \href{https://arxiv.org/abs/2211.09423}{DexPoint: Generalizable Point Cloud Reinforcement Learning for Sim-to-Real Dexterous Manipulation}.
\newblock \emph{Conference on Robot Learning (CoRL)}, 2022.

\bibitem[Christen et~al.(2022)Christen, Kocabas, Aksan, Hwangbo, Song, and Hilliges]{christen2022dgrasp}
S.~Christen, M.~Kocabas, E.~Aksan, J.~Hwangbo, J.~Song, and O.~Hilliges.
\newblock \href{https://ieeexplore.ieee.org/document/9880342}{D-grasp: Physically plausible dynamic grasp synthesis for hand-object interactions}.
\newblock In \emph{{Computer Vision and Pattern Recognition (CVPR)}}, 2022.

\bibitem[Wan et~al.(2023)Wan, Geng, Liu, Shan, Yang, Yi, and Wang]{Wan_2023_ICCV}
W.~Wan, H.~Geng, Y.~Liu, Z.~Shan, Y.~Yang, L.~Yi, and H.~Wang.
\newblock \href{https://ieeexplore.ieee.org/document/10378503}{{UniDexGrasp++}: Improving Dexterous Grasping Policy Learning via Geometry-Aware Curriculum and Iterative Generalist-Specialist Learning}.
\newblock In \emph{{International Conference on Computer Vision ({ICCV})}}, 2023.

\bibitem[Yuan et~al.(2024)Yuan, Zhou, Fu, and Lu]{yuan2024cross}
H.~Yuan, B.~Zhou, Y.~Fu, and Z.~Lu.
\newblock \href{https://arxiv.org/abs/2410.02479}{Cross-Embodiment Dexterous Grasping with Reinforcement Learning}.
\newblock \emph{arXiv preprint arXiv:2410.02479}, 2024.

\bibitem[Zhang et~al.(2024{\natexlab{a}})Zhang, Christen, Fan, Hilliges, and Song]{zhang2024graspxl}
H.~Zhang, S.~Christen, Z.~Fan, O.~Hilliges, and J.~Song.
\newblock \href{https://link.springer.com/chapter/10.1007/978-3-031-73347-5_22}{{GraspXL}: Generating Grasping Motions for Diverse Objects at Scale}.
\newblock In \emph{{European Conference on Computer Vision (ECCV)}}, 2024{\natexlab{a}}.

\bibitem[Zhang et~al.(2024{\natexlab{b}})Zhang, Christen, Fan, Zheng, Hwangbo, Song, and Hilliges]{zhang2024artigrasp}
H.~Zhang, S.~Christen, Z.~Fan, L.~Zheng, J.~Hwangbo, J.~Song, and O.~Hilliges.
\newblock \href{https://ieeexplore.ieee.org/document/10550470}{{ArtiGrasp}: Physically Plausible Synthesis of Bi-Manual Dexterous Grasping and Articulation}.
\newblock In \emph{International Conference on 3D Vision (3DV)}, 2024{\natexlab{b}}.

\bibitem[Agarwal et~al.(2023)Agarwal, Uppal, Shaw, and Pathak]{Agarwal2023functional}
A.~Agarwal, S.~Uppal, K.~Shaw, and D.~Pathak.
\newblock \href{https://arxiv.org/abs/2312.02975}{Dexterous Functional Grasping}.
\newblock In \emph{Conference on Robot Learning (CoRL)}, 2023.

\bibitem[Lum et~al.(2024)Lum, Matak, Makoviychuk, Handa, Allshire, Hermans, Ratliff, and Wyk]{lum2024dextrahg}
T.~G.~W. Lum, M.~Matak, V.~Makoviychuk, A.~Handa, A.~Allshire, T.~Hermans, N.~D. Ratliff, and K.~V. Wyk.
\newblock \href{https://openreview.net/forum?id=S2Jwb0i7HN&noteId=S2Jwb0i7HN}{Dextr{AH}-G: Pixels-to-Action Dexterous Arm-Hand Grasping with Geometric Fabrics}.
\newblock In \emph{Conference on Robot Learning (CoRL)}, 2024.

\bibitem[Singh et~al.(2024)Singh, Allshire, Handa, Ratliff, and Wyk]{singh2024dextrahrgbvisuomotorpoliciesgrasp}
R.~Singh, A.~Allshire, A.~Handa, N.~Ratliff, and K.~V. Wyk.
\newblock \href{https://arxiv.org/abs/2412.01791}{DextrAH-RGB: Visuomotor Policies to Grasp Anything with Dexterous Hands}.
\newblock \emph{arXiv preprint arXiv:2412.01791}, 2024.

\bibitem[{Universal Robots}()]{UR5}
{Universal Robots}.
\newblock \href{https://www.universal-robots.com/products/ur5-robot/}{UR5}.
\newblock \url{https://www.universal-robots.com/products/ur5-robot/}.

\bibitem[{Wonik Robotics}()]{Allegro}
{Wonik Robotics}.
\newblock \href{https://www.wonikrobotics.com/robot-hand}{Allegro robot hand}.
\newblock \url{https://www.wonikrobotics.com/robot-hand}.

\bibitem[Hwangbo et~al.(2018)Hwangbo, Lee, and Hutter]{hwangbo2018raisim}
J.~Hwangbo, J.~Lee, and M.~Hutter.
\newblock \href{https://ieeexplore.ieee.org/document/8255551}{Per-contact iteration method for solving contact dynamics}.
\newblock \emph{{Robotics and Automation Letters (RA-L)}}, 2018.

\bibitem[Calli et~al.(2015)Calli, Singh, Walsman, Srinivasa, Abbeel, and Dollar]{ycb}
B.~Calli, A.~Singh, A.~Walsman, S.~Srinivasa, P.~Abbeel, and A.~M. Dollar.
\newblock \href{https://ieeexplore.ieee.org/document/7251504}{The YCB object and Model set: Towards common benchmarks for manipulation research}.
\newblock \emph{{International Conference on Advanced Robotics (ICAR)}}, 2015.

\bibitem[Deitke et~al.(2022)Deitke, Schwenk, Salvador, Weihs, Michel, VanderBilt, Schmidt, Ehsani, Kembhavi, and Farhadi]{objaverse}
M.~Deitke, D.~Schwenk, J.~Salvador, L.~Weihs, O.~Michel, E.~VanderBilt, L.~Schmidt, K.~Ehsani, A.~Kembhavi, and A.~Farhadi.
\newblock \href{https://arxiv.org/abs/2212.08051}{Objaverse: A Universe of Annotated 3D Objects}.
\newblock \emph{arXiv preprint arXiv:2212.08051}, 2022.

\bibitem[Chang et~al.(2015)Chang, Funkhouser, Guibas, Hanrahan, Huang, Li, Savarese, Savva, Song, Su, Xiao, Yi, and Yu]{shapenet2015}
A.~X. Chang, T.~Funkhouser, L.~Guibas, P.~Hanrahan, Q.~Huang, Z.~Li, S.~Savarese, M.~Savva, S.~Song, H.~Su, J.~Xiao, L.~Yi, and F.~Yu.
\newblock \href{https://arxiv.org/abs/1512.03012}{ShapeNet: An Information-Rich 3D Model Repository}.
\newblock Technical Report arXiv:1512.03012, 2015.

\bibitem[Wen et~al.(2024)Wen, Yang, Kautz, and Birchfield]{wen2024foundationpose}
B.~Wen, W.~Yang, J.~Kautz, and S.~Birchfield.
\newblock \href{https://arxiv.org/abs/2312.08344}{{FoundationPose}: Unified 6D Pose Estimation and Tracking of Novel Objects}.
\newblock In \emph{{Computer Vision and Pattern Recognition (CVPR)}}, 2024.

\bibitem[Kirillov et~al.(2023)Kirillov, Mintun, Ravi, Mao, Rolland, Gustafson, Xiao, Whitehead, Berg, Lo, Dollár, and Girshick]{kirillov2023sam}
A.~Kirillov, E.~Mintun, N.~Ravi, H.~Mao, C.~Rolland, L.~Gustafson, T.~Xiao, S.~Whitehead, A.~C. Berg, W.-Y. Lo, P.~Dollár, and R.~Girshick.
\newblock \href{https://arxiv.org/abs/2304.02643}{Segment Anything}.
\newblock \emph{arXiv preprint arXiv:2304.02643}, 2023.

\bibitem[Bai et~al.(2023)Bai, Bai, Yang, Wang, Tan, Wang, Lin, Zhou, and Zhou]{bai2023qwen}
J.~Bai, S.~Bai, S.~Yang, S.~Wang, S.~Tan, P.~Wang, J.~Lin, C.~Zhou, and J.~Zhou.
\newblock \href{https://arxiv.org/abs/2308.12966}{{Qwen-VL}: A Versatile Vision-Language Model for Understanding, Localization, Text Reading, and Beyond}, 2023.

\bibitem[Wen and Bekris(2021)]{wen2021bundletrack}
B.~Wen and K.~E. Bekris.
\newblock \href{https://dl.acm.org/doi/10.1109/IROS51168.2021.9635991}{{BundleTrack}: 6D Pose Tracking for Novel Objects without Instance or Category-Level 3D Models}.
\newblock In \emph{IEEE/RSJ International Conference on Intelligent Robots and Systems}, 2021.

\bibitem[Jonathan()]{iksolver}
F.~Jonathan.
\newblock \href{https://github.com/fjonath1/python_UR5_ikSolver}{python\_UR5\_ikSolver}.

\end{thebibliography}

\clearpage
\appendix

\section{Experiment Details}
\subsection{Large Scale Object Details}
In simulation, we utilize the Objaverse \cite{objaverse} objects processed to graspable sizes by \cite{zhang2024graspxl}, which contain different scales: small, medium, and large. Furthermore, as \cite{zhang2024graspxl} focuses on grasping motion generation without hand-table collision, we filter out the objects that are not feasible to be grasped from the table (diameter \scalebox{0.8}{$<$} 1cm, height \scalebox{0.8}{$<$} 2cm, or width \scalebox{0.8}{$>$} 15cm), leading to 247,786 objects in total.

\begin{figure}[!h]
	\centering
	\includegraphics[width=0.99\columnwidth]{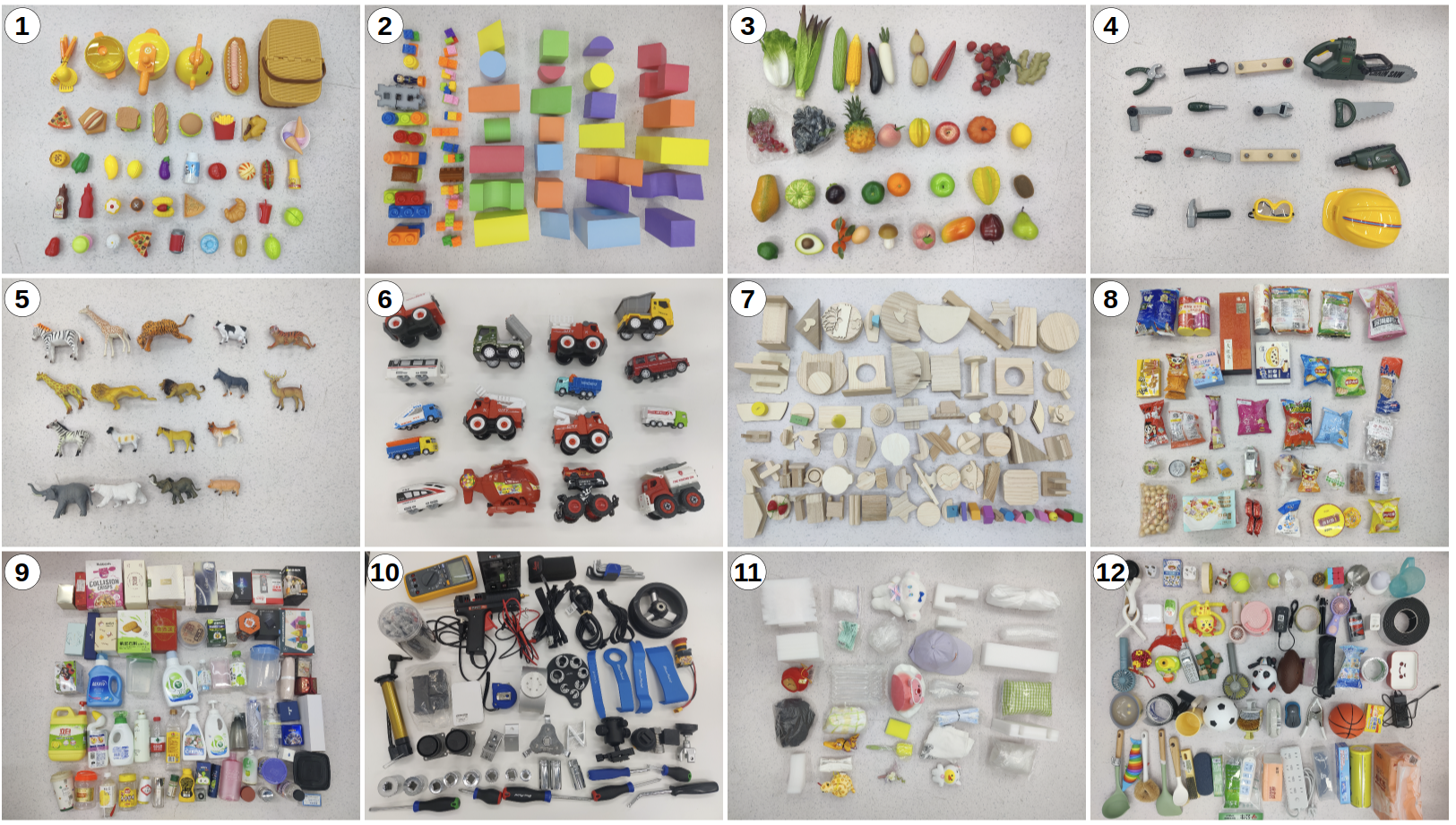}
	\caption{Real objects used for large-scale evaluation}
	\label{fig:objects}
\end{figure}

For hardware experiments, we choose 512 objects from 12 categories as explained in the main paper, which contains objects with various shapes, materials, masses, and sizes. The objects are visualized in \figref{fig:objects}, and the physical attributes of each category are listed in \tabref{tab:physical}. 

\begin{table*}[!h]
    \caption{Physical Attributes of Real Objects}
    \centering
       \resizebox{0.99\linewidth}{!}{
    \begin{tabular}{c|c|c|c|c}
        \toprule
        Category & Num. & Mass (g) & Scale (mm) & Material  \\
        \midrule
        Picnic Models            & 41 & 9-257  & 50*50*40 - 280*170*100 & Plastic \\
        Building Blocks          & 54 & 10-180 & 40*30*30 - 200*200*100  & Plastic/Styrofoam \\
        Fruit \& Vegetable Models& 35 & 7-196  & 70*50*50 - 330*80*80  & Plastic/Styrofoam/Rubber \\
        Tool Models              & 16 & 20-270 & 70*40*15 - 400*140*130  & Plastic \\
        Animal Models            & 18 & 26-165 & 100*50*30 - 230*120*100  & Rubber \\ 
        Toy Cars                 & 16 & 40-117 & 90*35*30 - 110*100*70  & Plastic \\ 
        Wooden Models            & 78 & 25-218 & 50*30*30 - 170*90*90  & Wood\\
        Snacks                   & 38 & 22-570 & 60*40*40 - 350*130*70  & - \\
        Bottles \& Boxes         & 67 & 15-550 &  35*35*30 - 240*170*55 & Plastic/Glass/Paper \\
        Real Tools               & 50 & 16-610 & 40*40*30 - 130*100*90  & Metal/Plastic/Rubber \\
        Deformable Objects       & 31 & 8-142  & 70*50*50 - 220*180*90  & Rubber/Cloth/Sponge/Styrofoam \\
        Other Daily Objects      & 68 & 19-454 & 40*30*30 - 270*130*100  & - \\
        \midrule
        Total & 512 & 7-610 & 35*30*15 - 400*200*130 & - \\
        \bottomrule
    \end{tabular}}
    \label{tab:physical}
\end{table*}

\subsection{Baseline Details}

\begin{figure}[h]
	\centering
	\includegraphics[width=0.6\columnwidth]{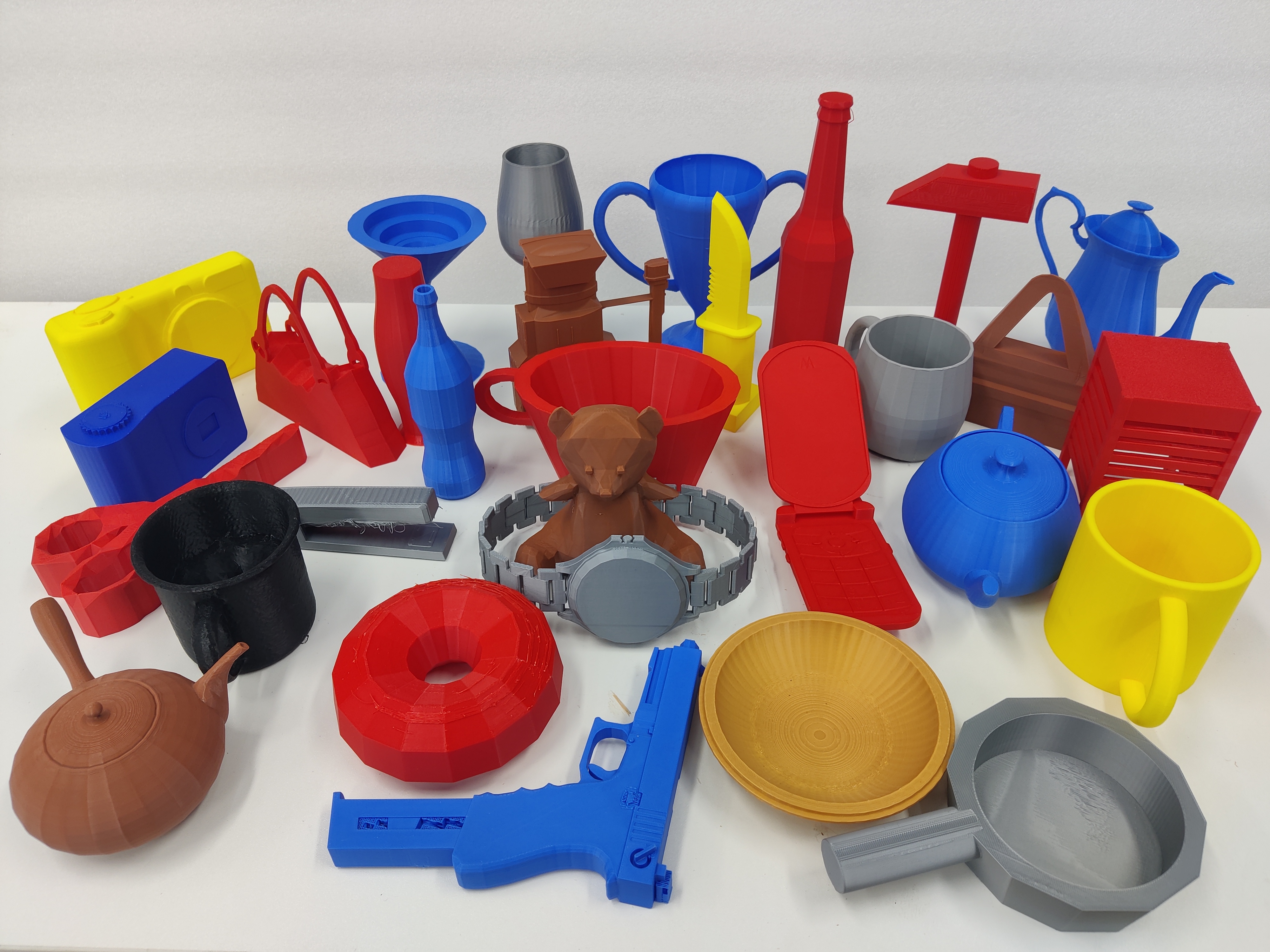}
	\caption{Objects used for comparisons and ablation}
	\label{fig:robustness_objects}
\end{figure}

\textbf{DexGraspNet}.
Since DexGraspNet optimizes grasping poses without considering tables, we incorporate a table collision loss to generate collision-free poses for a fair comparison. Specifically, given the object pose and corresponding table surface with height $h_{table}$, we add a loss term with the formulation $l_{table} = \sum_{i=0}^N ||h_{table} - h_i||^2 \cdot \mathcal{I}_{h_{table} > h_i}$, where $h_i$ is the height of the $i_{th}$ hand joints and N is the number of joints (including virtual joints for fingertips). To ensure a fair comparison, the initial hand pose for DexGraspNet is set identically to that used in our method.
As DexGraspNet leverages full object point clouds, we provide such inputs by using known object meshes and pose estimations from FoundationPose~\cite{wen2024foundationpose} during its hardware experiments. In contrast, our method is evaluated with our original setting using only single-view point clouds without access to object meshes. The 3D-printed ShapeNet~\cite{shapenet2015} objects used for the evaluation are shown in \figref{fig:robustness_objects}.

\begin{figure}[!h]
	\centering
	\includegraphics[width=0.7\columnwidth]{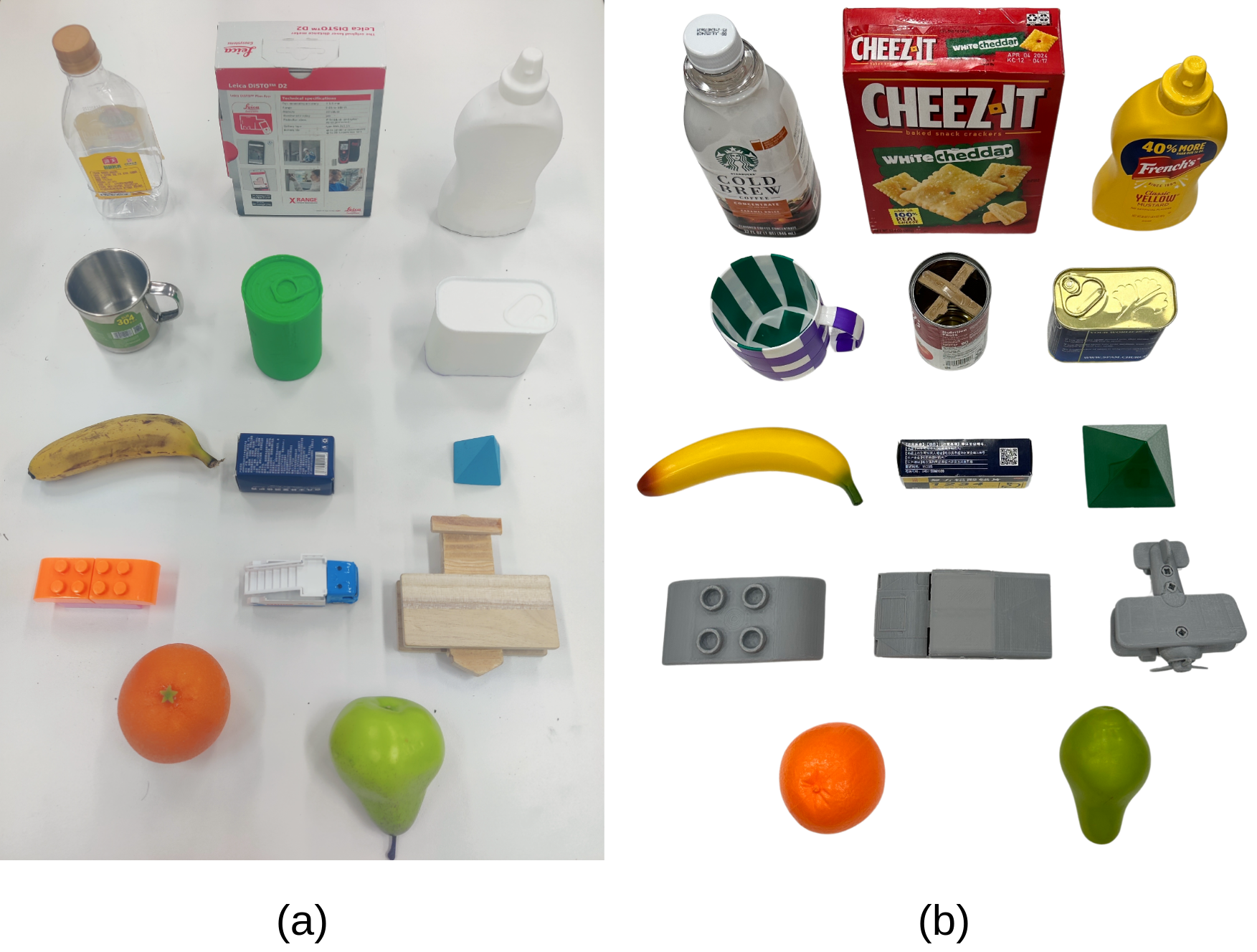}
	\caption{Objects used for comparison with SpringGrasp \cite{chen2024springgrasp} (a) and in their original experiment (b).}
	\label{fig:spring}
\end{figure}

\textbf{SpringGrasp}.
Since the exact objects used in SpringGrasp~\cite{chen2024springgrasp} are unavailable, we instead employ highly similar alternatives, as illustrated in \figref{fig:spring}. Given that both our method and SpringGrasp rely solely on depth information without RGB input, differences in object textures have minimal impact on performance. In fact, the objects used in our evaluation may pose greater challenges due to reflective surfaces or transparent materials, which introduce additional noise in the depth observations.
Focusing on robust grasping, we compare the success rates based on strictly successful grasps. Following a consistent evaluation criterion, we label partially successful grasps defined in SpringGrasp (objects are lifted but subsequently slip or slide) as failures for both methods.

\section{Implementation Details}

\subsection{Initial Pose}
\label{sec:pre-grasp}
We propose a simple yet effective method to initialize the arm and hand joints (see \figref{fig:hardware}), which are easily reachable via inverse kinematics (IK) planning since the hand does not interact with the object during initialization. The hand is initialized with partially open fingers using fixed joint angles $\textbf{q}_0=$ [0.2, 0.6, 0.2, 0.5, 0.2, 0.6, 0.2, 0.5, 0.2, 0.6, 0.2, 0.5, 1.3, 0.0, -0.1, 0.2].
To initialize the arm pose, we define a heading direction $\textbf{x}$ orthogonal to the palm (red arrow), pointing from the camera toward the object point cloud center $\textbf{c}$. This allows the hand to approach the object from a direction with minimal shape uncertainty. Next, we define a palm direction $\textbf{y}$ (green arrow) aligned with the final link of the arm. It is set to be orthogonal to $\textbf{x}$ while minimizing the projection length of the object point cloud along $\textbf{y}$. This helps the hand enclose the object from a narrow edge, which facilitates easier grasping.
The 6D wrist pose is finally defined by $\textbf{x}$, $\textbf{y}$, and an offset 25cm from $\textbf{c}$ along $\textbf{x}$, which is used to initialize the arm joints by an Inverse Kinematics (IK) solver~\cite{iksolver}.

\subsection{Object Point Cloud}
To extract object point clouds from a continuous visual stream, we first segment the object before grasping using SAM \cite{kirillov2023sam}, obtaining an initial, unoccluded point cloud. During grasping, we estimate the real-time point cloud under hand occlusion by tracking object pose changes with Cutie \cite{cheng2023putting} and BundleTrack \cite{wen2021bundletrack}. To mitigate depth noise, particularly for transparent or reflective objects, we apply outlier removal and per-frame smoothing to the initial point cloud over the first 50 frames.

\subsection{Action Space}
During grasping, our policy predicts both arm and hand target joint angles as actions. When the objects are grasped during deployment, we set fixed target angles for the arm joints to lift the objects.

\subsection{Domain Randomization}
As explained in the main paper, we randomize the environment parameters when training the student policy for robust sim-to-real transfer. The randomized parameters are listed in \tabref{tab:randomization}.

\begin{table}[!h]
    \caption{Domain randomization parameters.}
    \centering
       \resizebox{0.55\columnwidth}{!}{
    \begin{tabular}{c|c}
        \toprule
        Variable & Randomization  \\
        \midrule
        Friction Coefficient & \{0.5, 0.6, 0.7, 0.8,0.9\} \\
        Hand P Gain  & [0.9, 1.1] * 600 \\
        Hand D Gain  & [0.9, 1.1] * 20 \\
        Arm P Gain  & [0.5, 1.05] * 1.6e4 \\
        Arm D Gain  & [0.5, 1.05] * 600 \\
        Hand Joint Angles  & [-0.02, +0.02]rad + GT \\
        Arm Joint Angles  & [-0.005, +0.005]rad + GT \\
        Arm/Hand Link Position  & [-0.01, +0.01]m + GT \\
        Arm/Hand Link Orientation  & [-0.02, +0.02]rad + GT \\
        \bottomrule
    \end{tabular}}
    \label{tab:randomization}
\end{table}

\subsection{Training Details}
Using a single NVIDIA RTX 3090 GPU and 12 CPU cores, the training of the teacher and student policies takes approximately 30 hours in total. 
An overview of important parameters and reward function weights are provided in \tabref{tab:params} and \tabref{tab:weights}.

\begin{table}[!h]
\caption{Hyperparameters}
\centering
\resizebox{0.4\columnwidth}{!}{
\begin{tabular}{l|l}
\toprule
Hyperparameters PPO & Value \\
\midrule
Epochs & 1.5e4\\
Steps per epoch & 70\\
Environment steps per episode & 63 \\
Batch size & 2000 \\
Updates per epoch & 20 \\
Simulation timestep & 0.01s \\
Simulation steps per action & 20 \\
Discount factor $\gamma$ & 0.996 \\
Max. gradient norm & 0.5 \\
Value loss coefficient & 0.5\\
Entropy coefficient & 0.0\\
Hidden units & 128 \\
Hidden layers & 2 \\
\bottomrule
\end{tabular}
}
\label{tab:params}
\end{table}

\begin{table}[!h]
    \caption{Weights of the Reward Function}
    \centering
       \resizebox{0.5\columnwidth}{!}{
\begin{tabular}{l|l}
\toprule
Weights & Value \\
\midrule
$w^{d}$ (fingertip) & 2.0\\
$w^{d}$ (the other hand links) & 0.5\\
$w^{cd}$ (fingertip) & 6.0 \\
$w^{cd}$ (the other hand links)& 1.5\\
$w^{fd}$ (fingertip) & 4.0\\
$w^{fd}$ (the other hand links)& 1.0\\
$w^{cu}$ & -1.0\\
$w^{fu}$  & -0.5\\
$w^{h}$   & -0.05\\
$w_{h}$   & -1.0\\
$w_{o}$   & -15.0\\
$w_{l}$   & -5.0\\
$w_{q}$   & -1.0\\
$w_{re}$  & 1.0\\
$w_{act}$ & 1.0\\
$\lambda$ & 1.0 - $iter\_num$/2000 \\
\bottomrule
    \end{tabular}}
    \label{tab:weights}
\end{table}

\end{document}